\documentclass[10pt, a4paper]{article}
\usepackage{lrec}
\usepackage{graphicx}
\usepackage{tabularx}
\usepackage{soul}
\usepackage{amsmath}

\usepackage{epstopdf}
\usepackage[latin1]{inputenc}

\usepackage{hyperref}
\usepackage{xstring}

\title{\textbf{Translating Web Search Queries into Natural Language Questions}}

\name{Adarsh Kumar, Sandipan Dandapat, Sushil Chordia}

\address{AI \& Research, Microsoft \\
         Hyderabad, India \\
         \{adkuma, sadandap , sushilc\}@microsoft.com\\}

\abstract{
Users often query a search engine with a specific question in mind and often these queries are keywords or sub-sentential fragments.
In this paper, we are proposing a method to generate well-formed natural language question from a given keyword-based query, which has the same question intent as the query.Conversion of keyword based web query into a well formed question has lots of applications 
in search engines, Community Question Answering (CQA) website and bots communication.
We found a synergy between query-to-question problem with standard machine translation (MT) task. We have used both Statistical MT (SMT) and Neural MT (NMT) models to generate the questions from query. We have observed that MT models performs well in terms of both automatic and human evaluation.\\ \newline  }

\begin{document}

\maketitleabstract

\section{Introduction}
Search engines have improved a lot in last decade in all aspects. Earlier, the primary task of a search engine was to extract most relevant links for the query and present them as results. Lately, instead of just giving relevant links related to the query, search engines are trying to directly answer to any question asked. For example, for the query \textit{``japan's capital"} in modern search engines (eg. Bing and Google) directly answer \textit{``Tokyo"}, instead of providing a link containing the answer. Thus, search engines are evolving to save time for users and increase their productivity.
To further enhance the user-experience and increase productivity, search engines apart from showing the answer for a particular question, are trying to show related questions, to help users in their exploration. For example, for the query \textit{``fever symptoms''}, user mostly wants answer to the question \textit{``What are the symptoms of fever?"} and for the same query, questions like \textit{``How do you treat fever?"}, \textit{``What causes high fever?''} are highly related. 
To show related questions, search engines need to have a well framed question corpus from which they can extract relevant questions given a query.
\cite{white2015questions} have shown that more than 10\% of queries issued on a search engine has question intent whereas only 3\% of them are formulated as natural language questions. Most of these queries are primarily keywords or sentence fragments.
Hence, a corpus of questions can not be created directly using the search queries with question intent due to the issue of grammatical correctness and incomplete sentence formation.
To overcome this problem, we are proposing a technique to convert query with question intent, into a well-formed question. This technique can be used to generate well formed questions asked by the user, which can be used by search engines.
Apart from the direct application in search engines, query keywords to question conversion has applications in Question Answering (QA) systems, bots communication, Community Question Answer (CQA) websites etc. In CQA websites, when users have typed some keywords to search for questions, one can generate the questions and help them in framing the question using question corpus. Digital assistants can use this technology to refine the intent of query in natural language and help navigate the user to his/her exact needs. 

Query to question conversion was first suggested by \cite{lin2008automatic}, where he pointed out it's application in CQA websites and richer query expansion. 
Lin's idea was further extended by \cite{zhao2011automatically}, in which they have followed a template-based approach. They generate templates from $\langle query,question\rangle$ pairs from search logs and CQA websites and instantiate the template on the input query. 
At the same time, \cite{zheng2011k2q} also used a similar template-based technique. They generate templates from the question collected from CQA websites. They used a single variable templates, which essentially replaced a single word by some placeholder. Thus, the framework heavily relies on existing questions.
Another similar work was done by \cite{kalady2010natural} in which they derived question from a well formed sentence using parse tree and named entity recognitions. Their system is limited to certain types of questions.
Most of the techniques used to generate question from query are rule-based which are limited by the variety of question rules/templates, grammatical correctness, relevance between query and generated question etc. In this paper we propose a novel statistical approach to generate well-formed question from search keywords. The primary contribution of our work is that we have reduced the problem of query to question conversion into a translation problem. Furthermore, we also have shown how to build $\langle query, question \rangle$ parallel corpus from web search log that retain users' intention between query and question pair.Table \ref{tab:table1} shows some of the extracted pairs. We have made a detailed comparison between different translation framework with respect to our problem. 
\begin{table}\small
\begin{tabular}{ p{2.8cm}p{4.8cm}}
 \hline
 \textbf{Queries}     & \textbf{Questions} \\
 \hline
 fever symptoms   & What are the symptoms of fever ?    \\
 japan capital &   What is the capital of japan ?  \\
 string to int c\# & How to convert string to int in C\# ?  \\
 cancer types     & What are different types of cancer ? \\
 \hline
\end{tabular}
\caption {Example of queries and related questions}
\label{tab:table1}
\end{table}
\section{Approach}
The query to question generation problem can be formally stated as follows: given a sequence of query keywords $\textit{k}$ ($k_1, k_2, \ldots,k_n$) we want to generate the corresponding natural language question $\textit{q}$ ($q_1, q_2, \ldots, q_m$). This can be seen as a translation problem between source language sentence $k$ and target language sentence $q$. Note that both $k$ and $q$ are in English language while $q$ is a syntactically and semantically correct sentence of the language but $k$ is a grammatically ill-formed query.
In this work, we first use a SMT-based \cite{koehn2003statistical} approach. We have used the most widely used vanilla Moses\footnote{\url http://www.statmt.org/moses/} to build the SMT system. 
We consider this as the baseline system and call it \textbf{SMT}.
We use a NMT-based approach as described by~\cite{bahdanau2014neural}. Our NMT-based model uses bidirectional RNN with attention model~\cite{cho2014learning,sutskever2014sequence,schuster1997bidirectional}. Given an input sequence $\textit{k}$ from source language, i.e. queries, we want to generate a sequence $\textit{q}$ of target language, i.e. questions, which has similar question intent. We want to find the $\textit{q}$ which maximizes $arg\max_{\textit{q}}p(\textit{q}|\textit{k})$. We train a neural model which learns to maximize the conditional probability for sequence pairs in our parallel training corpus. After the model is trained, on giving a sequence $\textit{k}$ from source language, it generates a sequence $\textit{q}$ of target language which maximizes the conditional probability.

Our neural machine translation model consists of an encoder and a decoder. Encoder learns a fixed length representation for variable length input sequences and decoder takes that fixed length learned representation as input and generates the output sequence.
For example, for input sequence vectors $\textit{k}$ ($k_1, k_2, \ldots, k_n$), encoder encodes this into a fixed dimension vector $rep$. In general RNN's are used, such that :
\begin{equation}
h_{t} = f(k_{t},h_{t-1})
\end{equation}
\vspace{-10pt}
\begin{equation}
rep = z(h_{1},h_{2},...h_{T})
\end{equation}
$h_{t}$ is the hidden state at time $t$ and $k_{t}$ is input sequence at time $t$. $f$ and $q$ are non-linear functions. In our model we are using $f$ as LSTM \cite{hochreiter1997long} and define $z$ as in equation (3):
\begin{equation}
z(h_{1},h_{2},...,h_{t}) = h_{t}  
\end{equation}
The encoder tries to store the context of the input sequence into vector $rep$. During training, decoder learns to maximize the conditional probability. Decoder defines a conditional probability over the translation sequence $\textit{k}$ as follows : 
\begin{equation}
\begin{split}
p(\mathbf{\textit{q}}) & = \prod_{t=1}^{T}p(q_{t}|{q_{1},q_{2},...q_{t-1}},rep) \\ &= \prod_{t-1}^{T}g(q_{t-1},s_{t},rep)
\end{split}
\end{equation}
where $\mathbf{\textit{q}}$ = (${q_{1},q_{2},\ldots,q_{T}}$) and $g$ is non-linear. 
We are using attention model \cite{bahdanau2014neural}, in which conditional probability gets changed to following:
\begin{equation}
p(q_{i}|{q_{1},q_{2},\ldots,q_{i-1}}, \mathbf{\textit{k}}) = g(q_{i-1}, s_{i}, rep_{i})
\end{equation}
where $s_{i}$ is : 
\begin{equation}
s_{i} = g(q_{i}, s_{i-1}, rep_{i})
\end{equation}
The context vector $rep_{i}$ is computed as below : 
\begin{equation}
rep_{i} = \sum_{j=1}^{T_{x}}\alpha_{ij}h_{j}
\end{equation}
The weight $\alpha_{ij}$ of each annotation $h_{j}$ is computed by
\begin{equation}
\alpha_{ij} = \frac{exp(e_{ij})}{\sum_{m=1}^{T_{x}}exp(e_{im})}
\end{equation}
where 
\begin{equation}
e_{ij} = a(s_{i-1}, h_{j})
\end{equation}
This approach allows decoder to decide which part of input it wants to pay attention. We have used BiRNN, which has two function $\overrightarrow{f}$ and $\overleftarrow{f}$, where $\overrightarrow{f}$ reads the input sequence from $k_1$ to $k_T$ and produces forward hidden states$(h_{f_1},h_{f_2},\ldots,h_{f_T})$, i.e. in usual order, and the $\overleftarrow{f}$ reads in opposite direction, i.e. $k_T$ to $k_1$ and generates hidden backward vectors $(h_{b_1}, h_{b_2},\ldots, h_{b_T})$. At time $t$, we get the final hidden vector by concatenating forward as well as backward hidden vector at time $t$. This way BiRNN helps in storing the context of not only the preceding words but also the following words.

\section{Experimental Setup and Results}
First we conduct our baseline experiment using Moses SMT system to compare the results with our NMT-based model. The Moses SMT system uses KenLM~\cite{Heafield-estimate} as the default language model and MERT~\cite{och2003minimum} to reestimate the model parameters. We shall call it \textbf{SMT}.
In our particular NMT-based approach, we implemented a BiRNN model using LSTM with attention. We used 2 layered deep LSTMs with 512 cells at each layer. We kept the embedding dimension to be 300. Our input vocabulary size for both source and target language, i.e., queries and question had 150,000 words. We used stochastic gradient descent with initial learning rate of 0.5 and learning rate decay factor of 0.99. We kept batch size to be 128 and trained the model for a total of 6 epochs.
\subsection{Data Used}
In this case, parallel data refers to the ($k$,$q$) pair where $k$ is a query with question intent and $q$ is the corresponding natural language question with same question intent. 
We used Bing's web search logs to create our parallel data. Bing's Search Log stores 3 basic things : 
\begin{itemize}
\itemsep0em 
\item Queries ($k$) searched on bing
\item The URLs ($U$) which were shown for those queries in search result page 
\item URL ($u\in U$) which was clicked by the user for the respective query
\end{itemize}
We filtered all the queries ($k$), which landed on a CQA website, which contains some question ($q$) and its answer. We extracted the question ($q$) from that clicked CQA website and create the pair ($k,q$) for our dataset.  
Our hypothesis behind this was that after querying in any search engine, users click on those links which they find satisfactory and those queries ($k$) after which a user clicks on a website containing a question ($q$), can be assumed to have a question intent. 
To make sure the questions in our dataset are grammatically correct, we only considered reputed CQA websites like WikiAnswers,\footnote{\url https://answers.wikia.com/wiki/Wikianswers} Quora,\footnote{\url https://www.quora.com} and Yahoo Answers.\footnote{\url https://in.answers.yahoo.com/} The hypothesis being that moderators on these CQA websites are pretty strict in maintaining quality questions. 
We only kept ($k,q$) pairs in which query ($k$) had less than 10 words to avoid garbage queries.
We also made sure that we only select those $(k,q)$ pairs, in which question started with either a ``wh'' word or other question words (e.g. what, where, who, how, is, can, did, list, are etc.). 
After all this filtering, we were left with around 13 Million query-question pair $(k,q)$. We used randomly drawn 5000 sentences for test and development set (each 2500 sentences), disjoint from the training data.
We found around 50\% of the queries have less than 5 words.  The average length of the query and question are 5.6 and 8.5, respectively. Also, 85\% of the questions are of ``what (53\%)'', ``how(21\%)'',``is(6\%)'' and ``who(5\%)'' types. Fig. 1 plots the Query Length Distributions and Fig. 2 plots the percentage of different types of questions in our dataset. 
\begin{figure}%
    \centering
    {{\includegraphics[width=8cm]{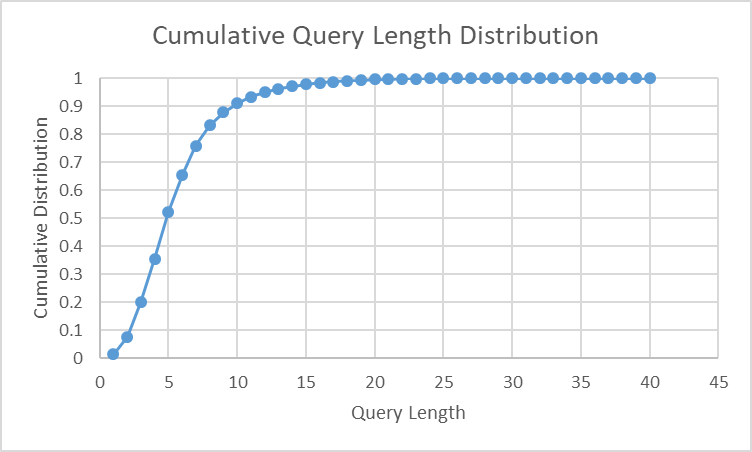} }}%
    \caption{Query Length Cumulative Distribution}
    \qquad
    {{\includegraphics[width=8cm]{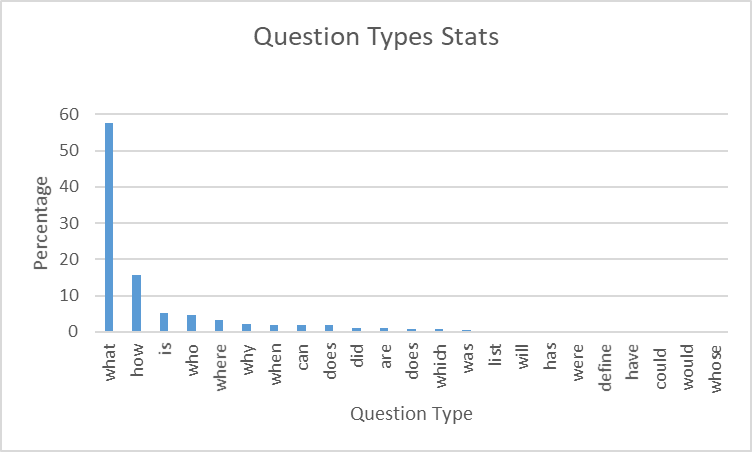} }}%
    \caption{Question Type Distribution}%
\end{figure}
\begin{table*}
\footnotesize
\begin{tabular}{p{2.5cm}p{4.3cm}p{4.3cm}p{4.3cm}}
\hline
  Query     & Generated Question by SMT   & Generated Question by NMT & Golden Truth\\
\hline
grams in 1 lb & how many grams are in 1 lb? & how many grams are in 1 pound? & how many grams are in 1 pound ?\\
anesthesiologist salary dubai & what is the salary of an anesthesiologist in dubai? & what is the salary of an anesthesiologist in dubai?&how much does an anesthesiologist make in dubai? \\
richest man in kansas & what is the richest men in kansas?  & who is the richest man in kansas? & Who is the most rich man of kansas?\\
small bone in human body located & what is the small bone in the body located? & where is the smallest bone in human body located?& where is the smallest bone in human body located?\\
first woman rapper & what was the first woman in the rapper & who was the first woman rapper?& who was the first woman rapper?\\
\hline
\end{tabular}
\caption{System Generated Output Produced by Different Models}
\label{tab:table_ex}
\end{table*}
\\
\subsection{Results}
In order to evaluate the performance of our system, we have used the most widely used MT evaluation metric BLEU~\cite{papineni2002bleu}. BLEU uses modified $n$-gram precision between the hypothesis and the reference. Note that the value of BLEU ranges from 0 to 100. 

First, in order to estimate the difficulty of the task we conducted an experiment (we shall call it \textbf{Identity Model}), we replicated input as the hypothesis translation, since both source query and target question are in English. This gives 19.33 BLEU score. This is due to large amount of vocabulary overlap between the query and its corresponding question.

The baseline SMT gives a BLEU score of 52.49 while NMT system has a BLEU score of 58.63. The NMT system has a 6.14 absolute BLEU point improvement compared to the SMT system. Both SMT and NMT system has a significant improvement over the identity model. The higher BLEU score ($>50$) by both SMT and NMT models are achieved due to the overlap between query and question keywords (as reflected in the BLEU score of the identity model). 
\subsection{Human Evaluation}
We conducted a human evaluation to judge the quality of the generated output. We manually evaluated approximately 1000 query/question pairs with the help of 12 people (more than 5 years of experience of using search engines). For each query-generated output pair, we asked participants following questions : 
\begin{itemize}
\itemsep0em
  \item Is the question grammatically correct?
  \item How similar is the intent between query and generated output?  
\end{itemize}
First question was a Yes-No based question and for the second question, participants were asked to judge the question intent similarity on a scale of $1-5$ between the pair, with $5$ being highly similar.
\begin{figure}
\includegraphics[width=\linewidth,height=4.8cm]{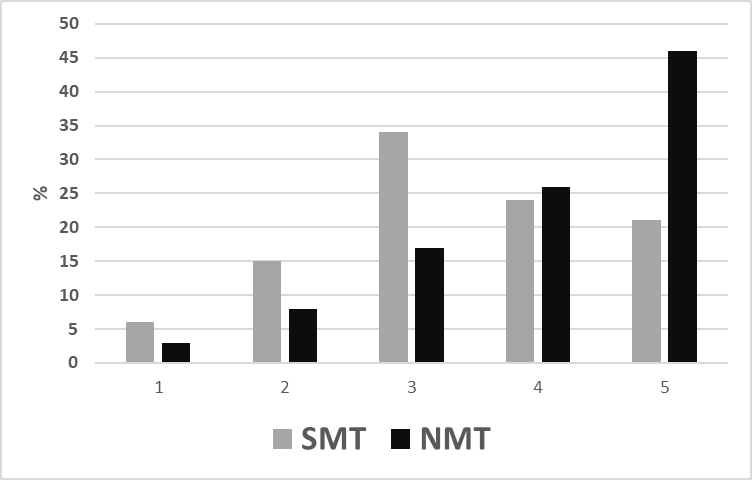} 
\caption{Intent Similarity Score Distribution}
\label{fig:NMTFig}
\end{figure}
In terms of grammatical correctness of the output generated from the two models, around 63\% of output generated from SMT were grammatically correct, while with NMT, almost 86\% of output were grammatically correct. SMT often make errors due to incorrect choice of question words as shown in examples in Table 2. SMT often choose ``what'' due to its high frequency in the corpus (cf. Section 3.1). 
In terms of intent similarity, around 72\% of the question generated by NMT model received very high score (4 and 5) in intent similarity by human evaluators, compared to only 45\% in case of SMT. Figure \ref{fig:NMTFig} shows the distribution of scores both model got from human evaluators.
We observed that NMT model performed better than baseline SMT in terms of BLEU score evaluation, as well as human based judgement. 
\section{Conclusions}
In this paper we have described machine-translation based approach for automatic generation of well-formed question from keyword-based query. We used automatically extracted parallel data from search logs to train the models. Our experiments shows that NMT models work better compared to the baseline statistical model. The present model generates the most likely question from a search query which has explicit question intent. For future works we wish to add text from Search Result Page also as input along with the raw query, with the assumption being that the given text will provide more contextual information about the query.   

\section{Bibliographical References}
\label{main:ref}

\bibliographystyle{lrec}
\bibliography{xample}


\end{document}